\documentclass[letterpaper]{article} 
\usepackage{aaai23}  
\usepackage{times}  
\usepackage{helvet}  
\usepackage{courier}  
\usepackage[hyphens]{url}  
\usepackage{graphicx} 
\usepackage{natbib}  
\usepackage{caption} 
\usepackage{algorithm}
\usepackage{algorithmic}
\usepackage{newfloat}
\usepackage{listings}
\usepackage{xcolor} 
\usepackage{amssymb}
\usepackage{graphicx}
\usepackage{subcaption}
\usepackage{amsmath}
\usepackage{booktabs}
\usepackage{hyperref}
\hypersetup{
	colorlinks=true,
	citecolor=cyan, 
}
\usepackage{tabularx}
\usepackage{multirow}
\usepackage{booktabs}
\usepackage{colortbl}
\DeclareCaptionStyle{ruled}{labelfont=normalfont,labelsep=colon,strut=off} 
\floatstyle{ruled}
\newfloat{listing}{tb}{lst}{}
\floatname{listing}{Listing}

\title{Soft Masked Mamba Diffusion Model for CT to MRI Conversion}
\author{
	Zhenbin Wang\textsuperscript{\rm 1},
	Lei Zhang\textsuperscript{\rm 1}\thanks{The corresponding author.},
	Lituan Wang\textsuperscript{\rm 1},
	Zhenwei Zhang\textsuperscript{\rm 1}
}
\affiliations{
    \textsuperscript{\rm 1}Sichuan University, Chengdu, China\\
    \textsl{wangzhenbin@stu.scu.edu.cn, leizhang@scu.edu.cn}
}
\usepackage{bibentry}
\begin{document}
\maketitle

\begin{abstract}
Magnetic Resonance Imaging (MRI) and Computed Tomography (CT) are the predominant modalities utilized in the field of medical imaging. 
Although MRI capture the complexity of anatomical structures with greater detail than CT, it entails a higher financial costs and requires longer image acquisition times. 
In this study, we aim to train latent diffusion model for CT to MRI conversion, replacing the commonly-used U-Net or Transformer backbone with a State-Space Model (SSM) called Mamba that operates on latent patches.
First, we noted critical oversights in the scan scheme of most Mamba-based vision methods, including inadequate attention to the spatial continuity of patch tokens and the lack of consideration for their varying importance to the target task.
Secondly, extending from this insight, we introduce Diffusion Mamba (DiffMa), employing soft masked to integrate Cross-Sequence Attention into Mamba and conducting selective scan in a spiral manner.
Lastly, 
extensive experiments demonstrate impressive performance by DiffMa in medical image generation tasks,
with notable advantages in input scaling efficiency over existing benchmark models. The code and models are available at \textit{\textcolor{red}{https://github.com/wongzbb/DiffMa-Diffusion-Mamba}}.
\end{abstract}

\section{Introduction}
Medical images obtained through a variety of imaging techniques provide complementary diagnostic information on bodily tissues. As a widely adopted imaging modality in medicine, Computed Tomography (CT) imaging is rapid, cost-effective, and provides excellent imaging effects on bones and calcified tissues, but demonstrates limited soft tissue contrast, particularly in the brain, neck, and pelvic regions. However, Magnetic Resonance Imaging (MRI) can achieve high-definition delineation of soft tissue, yet it requires lengthy imaging times and involves significant costs. Using image generation models to convert CT into MRI is a viable option to expand the range of diagnostic examinations without increasing costs.

Image generation models, Diffusion models~\cite{ho2020denoising, sohl2015deep} and Generative Adversarial Networks (GANs)~\cite{goodfellow2014generative} in particular, have been firmly established as commonly used methods in image processing~\cite{rombach2022high} and video analysis~\cite{videoworldsimulators2024, li2024endora}. Among recent methods, Diffusion is preferentially adopted due to its stability and superior control of conditions compared to GANs in synthetic images. Many of these models are constructed based on Latent Diffusion Models (LDM)~\cite{rombach2022high}, which are commonly based on Convolution Neural Networks (CNNs)~\cite{li2024u} or Vision Transformers (ViTs)~\cite{vaswani2017attention, dosovitskiy2020image, yao2023goal} backbone. CNNs are constrained by their limited receptive fields. In contrast, compared with CNNs, ViTs realize global receptive fields and dynamically predicted weighting parameters through the attention mechanism. However, the computational overhead increases quadratically with increasing input size.
To address this issue, several studies~\cite{liu2021swin, dong2022cswin} have enhanced the efficiency of ViTs by limiting the computational window or strip size. However, these improvements compromise the model's capability to establish long-distance dependencies, resulting in reduced accuracy.

Recently, State-Space Models (SSMs)~\cite{gu2021efficiently, mehta2022long, gu2022parameterization} have shown considerable potential for modeling long sequence with efficient linear computational complexity in Natural Language Processing (NLP), with significant efforts devoted to enhancing their robustness and efficiency. Among these methods, one particularly noteworthy is Mamba~\cite{gu2023mamba}, which introduces time-varying parameters into SSMs, substantially enhancing model efficiency via parallel processing.
Inspired by the success of Mamba in NLP, recent studies have explored its application to vision~\cite{zhu2024vision, liu2024vmamba, guo2024mambair}, videos~\cite{yang2024vivim, li2024videomamba}, graph~\cite{wang2024graph}, medical~\cite{ma2024u, ruan2024vm, wang2024weak}, point cloud~\cite{liang2024pointmamba}, multimodal~\cite{xie2024fusionmamba, zhao2024cobra}, and other fields to achieve linear complexity while retaining the global receptive field.
However, as Mamba is primarily intended for modeling 1D sequences, converting 2D patch embeddings into 1D sequences for unidirectional scan may disrupt the spatial integrity of the image. This spatial positional information is crucial, particularly in tasks such as MRI generation, where high levels of image detail are demanded.
Various 2D scan manners have been proposed in previous works, such as SS2D~\cite{liu2024vmamba}, ES2D~\cite{pei2024efficientvmamba}, and Local Scan~\cite{huang2024localmamba}, yet some of them neglect spatial continuity. 
In response, we introduce the newly Spiral-Scan Module, which applies Mamba to 2D images by incorporating a continuity-based inductive bias to maximize the preservation of structural information.

\begin{figure}[h]
	\centering
	\begin{subfigure}[]{0.22\textwidth}
		\includegraphics[width=0.9\textwidth]{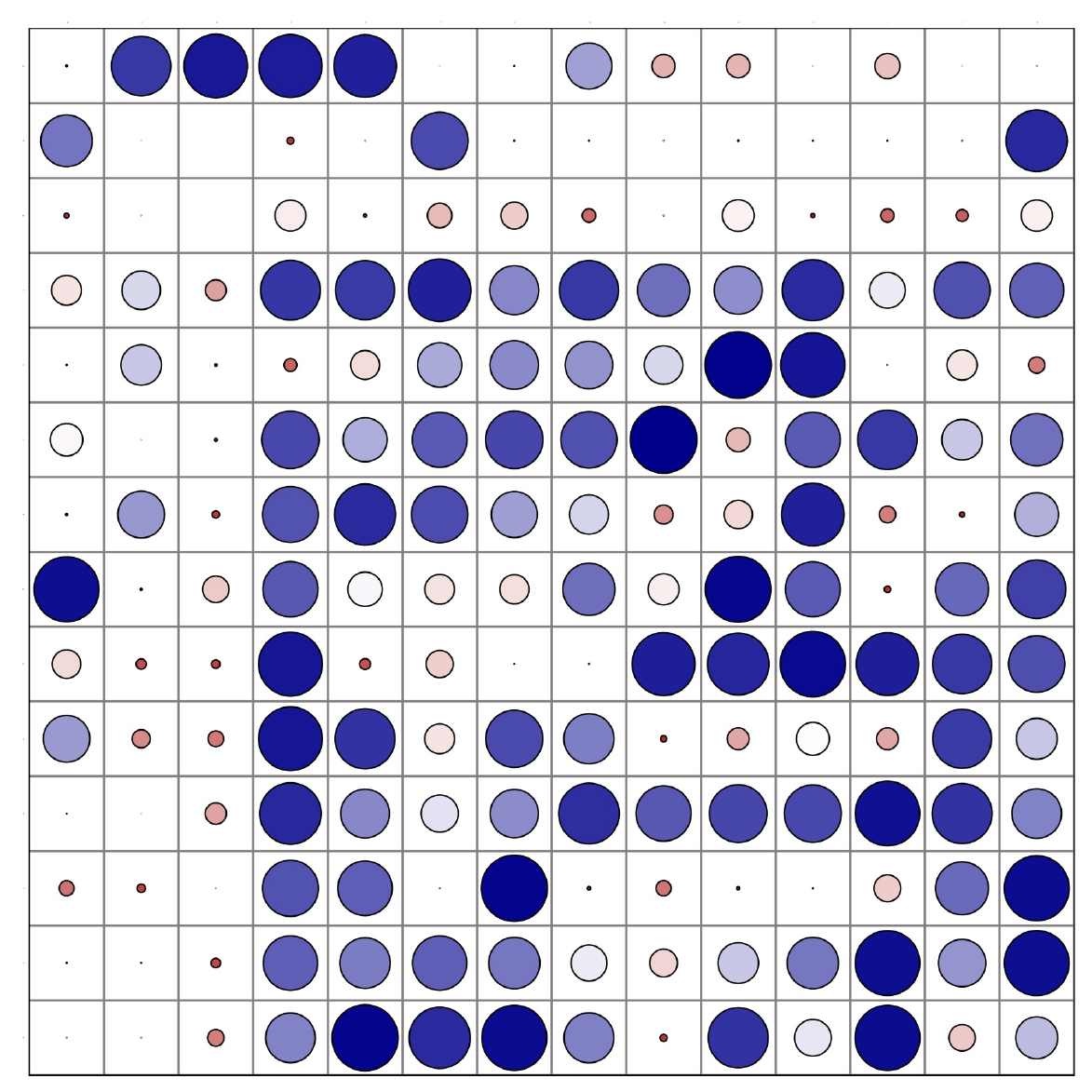}
	\end{subfigure}
	\begin{subfigure}[]{0.22\textwidth}
		\includegraphics[width=0.9\textwidth]{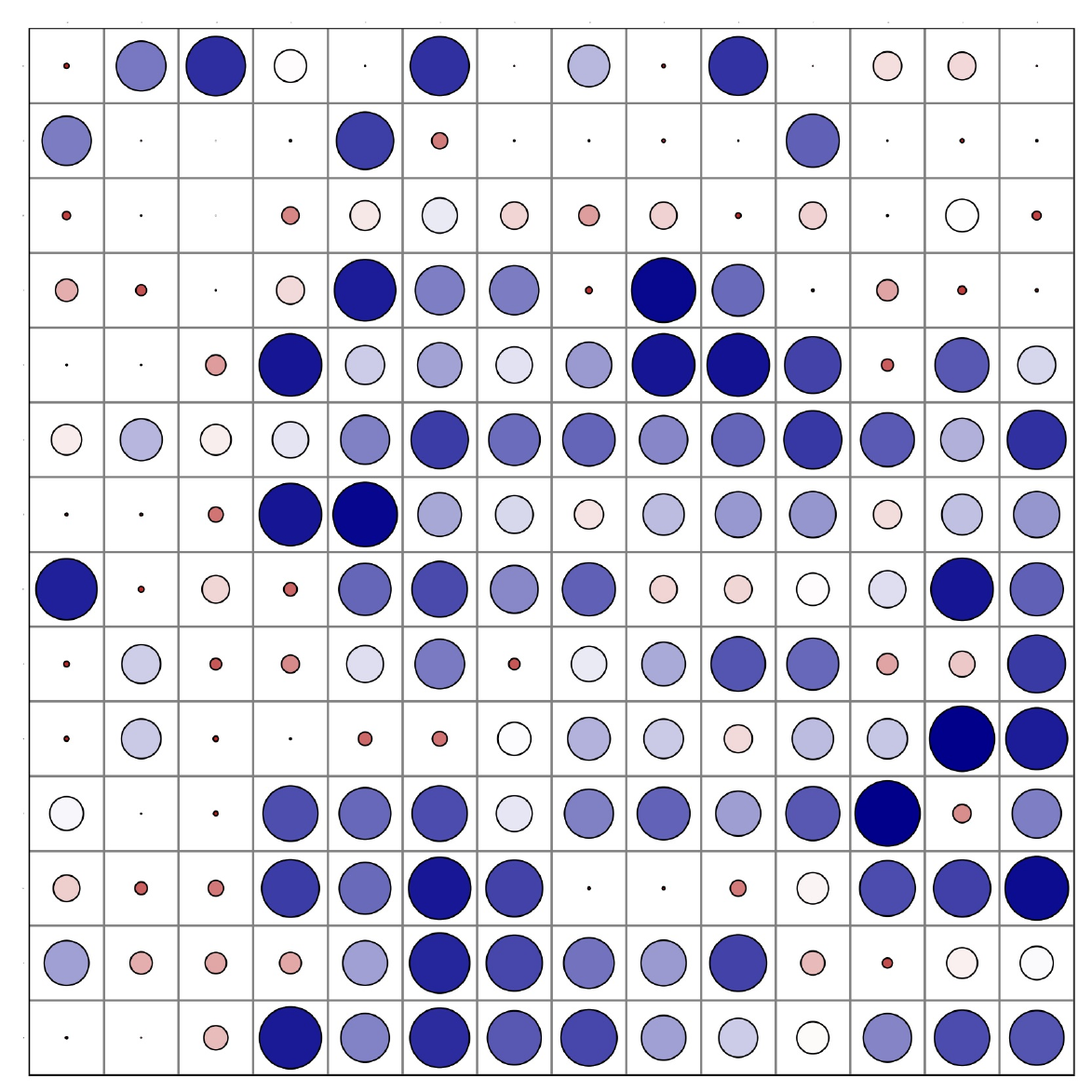}
	\end{subfigure}
	\caption{\textbf{Visualization of the significance of each 14x14 patch from two latent pelvic images}. The size and darkness of the circles denote the level of importance, with larger and darker circles indicating greater significance. The weights are derived from the pre-trained vision Embedder.}
	\label{fig:intro}
\end{figure}

To the best of our knowledge, previous work has overlooked a critical aspect where Mamba selective scan focuses solely on individual sequences and does not establish communication across sequences.
Unlike natural images, pathological tissues are typically concentrated in specific regions of a medical image. Consequently, the model is expected to prioritize these areas when performing the CT to MRI conversion task.
Specifically, we believe that different patch tokens within a sequence contribute variably to the generation of MRI images. By assigning higher weights to certain patches, the Mamba selective scan can more effectively focus on the most meaningful patches.
We propose improvements where, before training the diffusion model, a simple Vision Embedder is constructed and trained using cross-sequence supervision. This encoder is subsequently employed to generate token-level weights that are then integrated as conditions in the diffusion model, as shown in Figure~\ref{fig:intro}.

The main contributions of this study are summarized as follows:
\begin{itemize}
	\item We present a novel diffusion model framework that is exclusively based on Mamba, which we believe to be the first known method to use Mamba specifically for medical image generation tasks.
	\item Spiral-Scan is adopted to better handle the 2D spatial input, ensuring that the scanned sequence is spatially continuous, thereby preserving the contextual information of the image patches. 
	\item A soft mask module is proposed to integrate Cross-Sequence Attention into the Mamba selective scan mechanism, enhancing its ability to focus on tissue areas that are challenging to generate.
\end{itemize}

Extensive experiments were conducted on benchmark datasets to validate the effectiveness of the proposed method. With an equivalent parameter count, our DiffMa method achieves superior performance relative to traditional image generation methods based on CNNs and ViTs. Furthermore, we reimplement several existing visual Mamba block structures, and compared to them, our DiffMa method more effectively generates MRI images within the same number of iterations.

\section{Related Work}

\subsection{State-Space Models (SSMs)}
SSMs~\cite{gu2021efficiently, gu2021combining, smith2022simplified} represent a category of models that introduce state space transformations into deep learning. They encapsulate a sequence-to-sequence transformation and demonstrate their potential to handle tokens with long dependencies.
The state representation in early SSMs were computationally intensive and memory demanding, posing significant challenges during training.
\textit{Structured state space sequence} (S4)~\cite{gu2021efficiently} firstly proposed normalizing parameters into a diagonal structure. Subsequently, various advanced parameter normalization structures for SSMs have emerged, including decomposition of diagonal plus low-rank operations~\cite{hasani2022liquid}, multiple-input multiple output supporting~\cite{smith2022simplified}. \textit{Simplified State Space Layers for Sequence Modeling} (S5)~\cite{smith2022simplified} reduces the complexity to a linear level, and then \textit{Hungry Hungry Hippos} (H3)~\cite{fu2022hungry} incorporates it into the language modeling task.
\textit{Selective State Space Models} (S6)~\cite{gu2023mamba}, which builds on the foundation of S4 by enabling the model to selectively process information and concentrate on or disregard specific content, serves as an important component of Mamba.
These works effectively overcome the computational overhead problem of SSMs. 
With SSMs proving effective in NLP, efforts to include SSMs for computer vision has been growing. 
S4ND~\cite{nguyen2022s4nd} was the first to introduce S4-based SSMs to visual tasks, achieving performance comparable to ViTs. Following this, VMamba~\cite{liu2024vmamba} pioneered the application of S6-based SSMs to visual tasks. ZigMa~\cite{hu2024zigma} then merges Mamba and Diffusion for video and image generation tasks.

\subsection{SDE, SGMs and ODE in Diffusion Models}
Previous GANs-based image generation models attempt to reconstruct images directly from random white noise, which is susceptible to unstable imaging quality due to issues such as Mode Collapse~\cite{kodali2017convergence}. Diffusion models constitute a class of generative models grounded on probabilistic likelihood and derive their inspiration from non-equilibrium thermodynamics, often outperforming GANs in many cases. The fundamental concept involves using forward diffusion process to systematically perturb data distributions, followed by learning reverse diffusion process to restore these distributions.
Diffusion models involve several key components: Denoising Diffusion Probabilistic Models (DDPMs)~\cite{ho2020denoising}, Stochastic Differential Equations (SDEs)~\cite{song2020score}, and Score-based Generative Models (SGMs)~\cite{song2019generative}. The SDEs describe the process of noise addition in the Diffusion Models, and the SGMs enable the model to learn and reverse this noise addition process. The DDPMs integrates both SDEs and SGMs to form the complete architecture.
Recent studies~\cite{karras2022elucidating, lee2023minimizing} have replaced discrete SDEs with Ordinary Differential Equations (ODEs) to reduce sampling costs.
For established DDPMs, convolutional U-Nets or ViTs are the de-facto choice of backbone architecture; we explore pure Mamba.

\begin{figure*}[htbp]
	\centering
	\includegraphics[width=\textwidth]{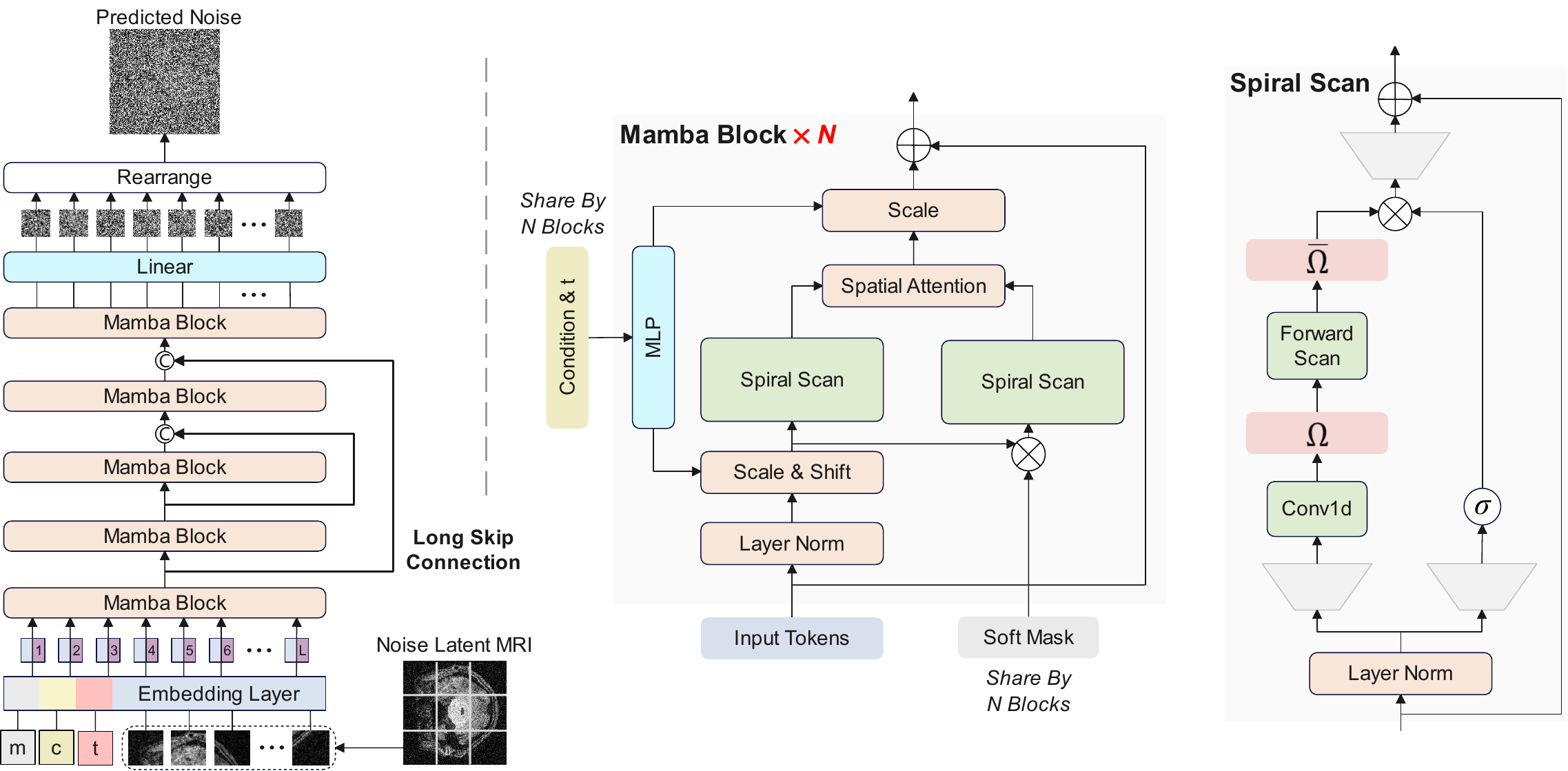}
	\caption{\textbf{The Diffusion Mamba (DiffMa) framework}. \textit{Left}: The overall framework of Diffusion. We use long skip connection to prevent Nan. \textit{Middle}: Details of Mamba blocks, consisting of two branches that use adaptive layer norms (AdaLN) to incorporate conditioning, and we introduce innovative soft mask to provide prior knowledge for sequences. \textit{Right}: Details of Mamba, where we employ Spiral-Scan to focus on the structural information.}
	\label{fig:framework}
\end{figure*}

\section{Preliminaries}
SSMs can be described as a class of time-invariant systems that transform a D-dimensional sequence $x(t)\in \mathbb{R}^{L \times D}$ into an output sequence $y(t)\in \mathbb{R}^{L \times D}$ via a learnable hidden state $h(t)\in \mathbb{R}^{N \times D}$. This process can be mathematically expressed as linear ODEs presented below:
\begin{equation} \label{eq1}
	\begin{split}
		&h'(t)=\mathbf{A} h(t)+\mathbf{B} x(t), \\
		&y(t)=\mathbf{C} h(t),
	\end{split}
\end{equation}
here $\mathbf{A}\in\mathbb{R}^{N \times N}$, $\mathbf{B},\mathbf{C}\in\mathbb{R}^{D \times N}$.

To facilitate calculations on a computer, it is necessary to discretize the preceding continuous differential equations. Prior studies~\cite{gu2021combining, gupta2022diagonal} have discretized the continuous parameters $(\mathbf{A},\mathbf{B})$ using zero-order hold (ZOH) method, considering a given sample timescale parameters $\Delta \in \mathbb{R}^D$, we have
\begin{equation} \label{eq2}
\begin{split}
	&\bar{\mathbf{A}} = e^{\Delta \mathbf{A}},\\
	&\bar{\mathbf{B}} = (e^{\Delta \mathbf{A}}-\mathbf{I})\mathbf{A}^{-1}\mathbf{B},\\
	&\bar{\mathbf{B}} \approx (\Delta \mathbf{A})(\Delta \mathbf{A})^{-1}\mathbf{A}\mathbf{B}=\Delta \mathbf{B},\\
	&\bar{\mathbf{C}} = \mathbf{C},
\end{split}
\end{equation}
where $\bar{\mathbf{A}} \in \mathbb{R}^{N \times N}$ and  $\bar{\mathbf{B}}, \bar{\mathbf{C}} \in \mathbb{R}^{N \times N}$. The approximation of $\bar{\mathbf{B}}$ in Eq.(\ref{eq2}) is derived using the first-order Taylor expansion.
Therefore, the discrete function of Eq.(\ref{eq1}) is
\begin{equation} \label{eq3}
	\begin{split}
		&h'(t)=\bar{\mathbf{A}} h(t-1)+\bar{\mathbf{B}} x(t), \\
		&y(t)=\bar{\mathbf{C}} h(t),
	\end{split}
\end{equation}
To simplify the computational process, the output is derived using the global convolution method, as illustrated below.
\begin{equation} \label{eq4}
	\begin{split}
		&\mathbf{y}=\mathbf{x}\circledast \bar{\mathbf{K}},\\
		\mathrm{with} \quad &\bar{\mathbf{K}} = (\bar{\mathbf{C}} \bar{\mathbf{B}},\bar{\mathbf{C}} \bar{\mathbf{A}}\bar{\mathbf{B}},\cdots,\bar{\mathbf{C}}\bar{\mathbf{A}}^{L-1}\bar{\mathbf{B}}).
	\end{split}
\end{equation}
Here $\bar{\mathbf{K}}\in \mathbb{R}^{L}$ is SSMs kernel, and $\circledast$ denotes convolution operation.
Mamba introduces a selection mechanism based on this concept, enabling continuous parameters $(\bar{\mathbf{B}}, \bar{\mathbf{C}}, \Delta)$ to vary with the input $\mathbf{x}$, thereby further extending the discretization process.
\begin{equation} \label{eq5}
	\begin{split}
		&\bar{\mathbf{B}}=f_{\bar{\mathbf{B}}}(\mathbf{x}),\\
		&\bar{\mathbf{C}}=f_{\bar{\mathbf{C}}}(\mathbf{x}),\\
		 &\Delta=softplus(\Delta+Broadcast(f_{\Delta}(\mathbf{x}))).
	\end{split}
\end{equation}
The $f_{\bar{\mathbf{B}}}(\cdot)$, $f_{\bar{\mathbf{C}}}(\cdot)$ and $f_{\Delta}(\mathbf{x})$ are linear projections. The choice of $Broadcast$ and $softplus$ is due to a connection to RNN gating mechanisms.

\section{Method}
In this section, we introduce the overall framework of our mehtod Diffusion Mamba (DiffMa), a new architecture for diffusion models. 
DiffMa specifically targets training diffusion models for MRI images, employing the visual Mamba to process sequences of patches.
Figure~\ref{fig:framework} illustrates the complete DiffMa framework. 
The individual components of DiffMa are delineated in detail below.
In particular, we concentrate on the \textit{Spiral-Scan} and \textit{Cross-Sequence Attention}. 

\subsection{Condition}
The input of DiffMa is noise latent space representation $z_{mri}$ generated by Variational Autoencoder (VAE) $f_{vae}$, for a $3 \times 224 \times 224$ MRI image $x_{mri}$, where the dimension of $z_{mri}$ is $4 \times 28 \times 28$. 
This $z_{mri}$ is first segmented into patches and then converted into $L$ tokens, each enhanced with positional embeddings. The size of $L$ is determined by the hyperparameters—patch size and strip size, and the dimension of each token is $D$. These tokens then serve as the input for a series of Mamba blocks, which process the information further.

Besides handling noise latent MRI images, these Mamba blocks also incorporate additional information, including the timesteps $t$, conditions $c$, and soft masks $m$. To facilitate the processing of CT images $x_{ct}$ within the Diffusion model, 
we employ two specialized encoders: the Vision Encoder $f_{clip}$ from BioMedCLIP~\cite{zhang2023biomedclip} and our pre-train Vision Embedder $f_{ve}$, both utilized to map CT images into embeddings, with their parameters fixed during the training of Diffusion model.
Among them, BioMedCLIP is a visual-language model trained with large-scale medical datasets, we employ its Visual Encoder to process the CT image $x_{ct}$. Concurrently, $f_{ve}$ is pre-trained by us, which takes as input the latent space representation $z_{ct}$ output by the VAE. The output from $f_{ve}$ includes not only the CT embedding $e_{ct}^{ve}$ but also soft mask $m$ that acts upon the sequence. The intuitive formulation is expressed as follows:
\begin{equation} \label{eq6}
	\begin{split}
		& e_{ct}^{clip} = f_{clip}(x_{ct}), \\
		& e_{ct}^{ve}, m = f_{ve}(f_{vae}(x_{ct})), \\
		& c = concate(e_{ct}^{ve}, mean(e_{ct}^{clip})),
	\end{split}
\end{equation}
where $e_{ct}^{clip}, t \in \mathbb{R}^{B \times D}$, $e_{ct}^{ve} \in \mathbb{R}^{B \times L \times D}$, $m \in \mathbb{R}^{B \times L}$, and $mean(\cdot)$ denotes the function calculating the mean across the middle dimension.

\subsection{Cross-Sequence Attention}
Cross-sequence attention is achieved by designing Vision Embedder to generate soft masks. Although S6 adopted by Mamba can selectively process information, it only scans a single sequence a posteriori and its selection principle is supervised by the loss function.
We believe that an excellent generative model should have the ability to distinguish subtle differences between its generated images. Therefore, we propose adding a branch to the Diffusion block to enhance this capability. 
Building on this perspective, we focus on tokens that cause significant differences in different sequences. 
Using MRI images to calculate soft masks directly is not feasible because MRI image priors are not available at the inference stage. Fortunately, CT image priors are available during both the inference and training stages.

\begin{figure}[htbp]
	\centering
	\includegraphics[width=0.4\textwidth]{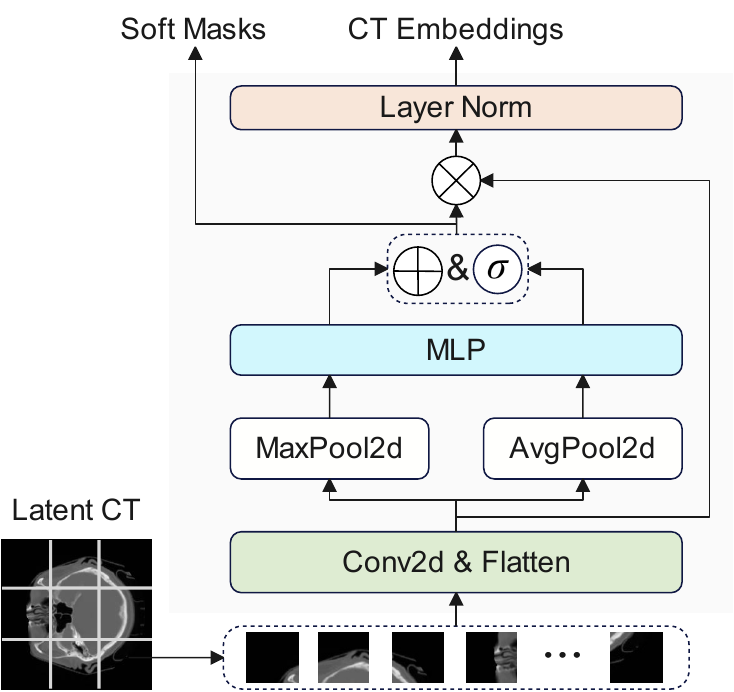}
	\caption{\textbf{Framework of Vision Embedder}. Unlike MRI patch tokens, CT patch tokens are trained without positional or temporal embeddings.}
	\label{fig:VisionEmb}
\end{figure}

The structure of our Vision Embedder $f_{ve}$ is depicted in Figure~\ref{fig:VisionEmb}. First, we divide the latent CT obtained by VAE into multiple patches, whose patch size is the same as the MRI in Figure~\ref{fig:framework}, so that the CT and MRI pixel areas are aligned. 
Then, we use 2D convolution maps these patches into embeddings with dimensions $B \times L \times D$, and use the channel attention mechanism to generate $B$ soft masks. These soft masks are also used to adjust the CT embeddings as part of the condition. 
We employ contrastive learning to train the Vision Embedding $f_{ve}$ to focus on the differences between CT images. Initially, tokens from each sequence are flattened into a 1D vector, denoted as $\bar{e}_{ct}^{ve} \in \mathbb{R}^{B \times (L \cdot D)}$, and then undergoes L2 normalization, where each row (each sample) is normalized independently. The similarity matrix $S$ is then calculated,
\begin{equation} \label{eq7}
	\begin{split}
		&\hat{e}_{ct,k}^{ve} = \frac{\bar{e}_{ct,k}^{ve}}{\left \| \bar{e}_{ct,k}^{ve} \right \|_2 }, \quad k=1,\cdots,B, \\
		& S = \frac{\hat{e}_{ct,k}^{ve} {(\hat{e}_{ct,k}^{ve}})^{\mathsf{T}}}{\tau  },
	\end{split}
\end{equation}
where $(\cdot)^{\mathsf{T}}$ denotes the transpose operation and $\tau$ is the temperature parameter. Then we calculate the cross InfoNCE loss~\cite{radford2021learning} $\mathcal{L}_{info}$ to train $f_{ve}$, the equation is as follows: 
\begin{equation} \label{eq8}
	\begin{split}
		&\mathcal{L}_{info}(f_{ve}) = -\frac{1}{B} \sum_{k=1}^{B}\log\frac{\exp(S_{i,i})}{\sum_{j=1}^{B}\exp(S_{k,j}) }  .
	\end{split}
\end{equation}

\subsection{Spiral-Scan}
\begin{figure}[htbp]
	\centering
	\includegraphics[width=0.45\textwidth]{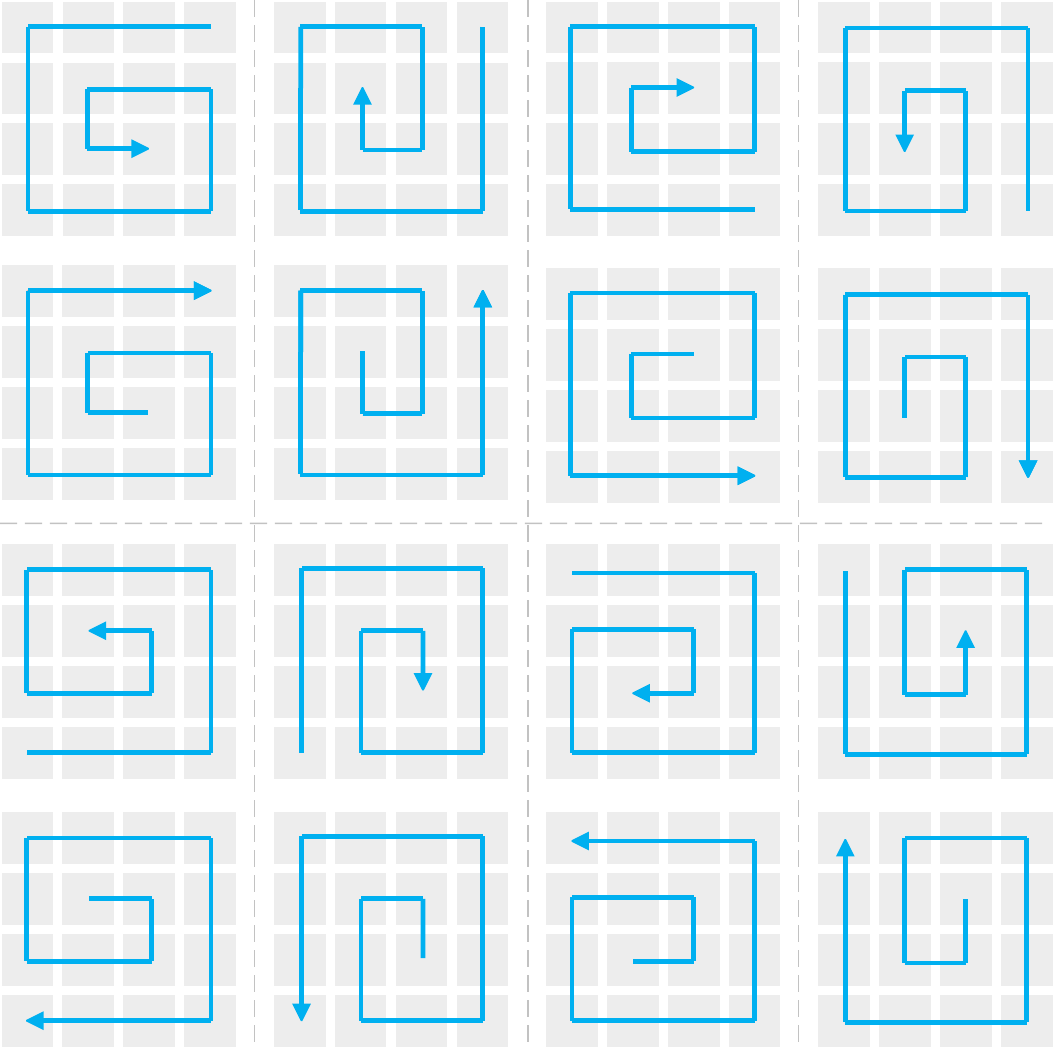}
	\caption{\textbf{The 2D Image Spiral-Scan}. There are eight schemes in total, each contains two modes, and every block employs one of these schemes.}
	\label{fig:scan}
\end{figure}

To ensure network awareness of position, as illustrated in the right part of Figure~\ref{fig:framework}, patch tokens of a sequence are initially rearranged via {\color{red} $\Omega$} before forward scan, and subsequently restored to their original order through {\color{red} $\bar{\Omega}$} after the scan is complete. 
Currently, related work in visual Mamba proposes various arrangement schemes to squeeze 2D patch tokens. However, these rearrangement schemes may not be optimal due to insufficient consideration of spatial continuity. 
To solve this problem, eight Spiral-Scan schemes have been designed, denoted as  $\Phi_{k}$ (where $k \in [0,7]$), as shown in Figure~\ref{fig:scan}, each scheme includes two modes in opposite directions. The scan scheme for the $i$-th Mamba block $\Omega_{i}=\Phi_{i\%8}$, where \% represents the modulo operation.

We combine the outputs of the two Spiral Scan blocks using the Spatial Attention Module~\cite{woo2018cbam}.
Diffusion models involve a series of steps that gradually add noise to the training data. Let $\beta_t$ denote the variance ratio of the noise added to the data at time $t$. Then, the noisy version of latent sample $z_{mri}$ at time $t$ can be expressed as $\sqrt{\bar{\alpha}_t}z_{mri} + \sqrt{1-\bar{\alpha}_t}\epsilon$, where $\epsilon$ is sampled from a Gaussian distribution with the same dimensions as $z_{mri}$, $\alpha_t=1-\beta_t$, and $\bar{\alpha}_t=\prod_{i=1}^{t}\alpha_i$. The The training loss for the denoising neural network $\epsilon_{\phi}(z_{mri}^{t};t)$ is
\begin{equation} \label{eq9}
	\begin{split}
		\mathcal{L}(\epsilon_{\phi}) \!=\! \mathbb{E}_{t,m,c} \| \epsilon_{\phi }(\!\sqrt{\bar{\alpha}_t}z_{mri} \!+\!\! \sqrt{1\!-\!\bar{\alpha}_t}\epsilon,t,c,m)\!-\!\epsilon  \|^2.
	\end{split}
\end{equation}

\subsection{Computation Analysis}
For a visual sequence of dimension $\mathbb{R}^{1 \times L \times D}$, $L$ is the number of tokens, i.e., the number of patches, and $D$ is the dimension of each token.
Regarding the computational complexity, global self-attention and our Sprial-Scan block are as follows:
\begin{equation} \label{eq8}
	\begin{split}
		&O(\mathrm{self\mathrm{-}attention}) = 4LD^2+2L^2D, \\
		&O(\mathrm{Spiral\mathrm{-}Scan}) = 2 \times [3L(2D)N + L(2D)N^2],
	\end{split}
\end{equation}
here $N$ is a fixed parameter, set to 16 by default. It is evident that the self-attention mechanism adopted by the Transformer exhibits quadratic complexity, while the Sprial-Scan block exhibits linear complexity.

\subsection{Model size}
We implement a sequence of $N$ DiffMa blocks, each operating with a hidden dimension of $D=512$, matching the image embedding dimension of BioMedCLIP. 
We utilize four distinct configurations: DiffMa-S, DiffMa-B, DiffMa-L, DiffMa-XL and DiffMa-XXL.
These configurations cover a comprehensive range of parameter size and FLOP allocations, ranging from 0.05 to 5.92 Gflops, providing the versatility to select the appropriate model size.
Detailed specifications of these configurations are presented in Table~\ref{table:config}.

\begin{table}[] 
	\centering
	\caption{Details of DiffMa models. The dimension of tokens is 512.}
	\begin{tabular}{@{}lcccc@{}}  
		\toprule
		Model  & Layers & Patch size & \#Params & FLOPs \\ \midrule
		DiffMa-S  & 4        & 7      & 10.16M    & 0.05G \\
		DiffMa-S  & 4        & 4      & 9.95M    & 0.17G \\
		DiffMa-S  & 4        & 2      & 9.88M    & 0.43G \\
		DiffMa-B  & 8        & 7      & 18.57M   & 0.09G \\
		DiffMa-B  & 8        & 4      & 18.37M   & 0.28G \\
		DiffMa-B  & 8        & 2      & 18.29M   & 0.85G \\
		DiffMa-L  & 16       & 7      & 35.39M   & 0.17G \\
		DiffMa-L  & 16       & 4      & 35.19M   & 0.45G \\
		DiffMa-L  & 16       & 2      & 35.11M   & 1.70G \\
		DiffMa-XL & 28       & 7      & 60.61M   & 0.29G  \\
		DiffMa-XL & 28       & 4      & 60.42M   & 0.78G  \\
		DiffMa-XL & 28       & 2      & 60.35M   & 2.96G  \\
		DiffMa-XXL & 56       & 7      & 119.50M   & 0.57G  \\
		DiffMa-XXL & 56       & 4      & 119.22M   & 1.55G  \\
		DiffMa-XXL & 56       & 2      & 119.30M   & 5.92G  \\
		\bottomrule
	\end{tabular}
	\label{table:config}
\end{table}

\begin{table*}[bp]
	\centering
	\caption{Performance comparison of CT to MRI conversion task on Pelvis and Brain. PSNR (dB) and SSIM (\%) are listed across the test set. \textit{To ensure the same number of model parameters, we employed the DiffMa-B model, which consists of 8-layers, while all other Mamba-base models use 13-layers. DiT use 7-layers. For ZigMa, we did not adopt the optional cross-attention. All models were trained for 50 epochs with batch size of 1}.}
	\label{table:compare}
	\begin{tabularx}{ 0.62\textwidth}{@{}lccc|ccc@{}}
		\toprule
		& \multicolumn{3}{c}{Pelvis} & \multicolumn{3}{c}{Brain} \\
		\midrule 
		& SSIM$\uparrow$ & PSNR$\uparrow$ & MSE$\downarrow$ & SSIM$\uparrow$ & PSNR$\uparrow$ & MSE$\downarrow$ \\
		LDM  \textit{{\scriptsize (CVPR'22)}}&40.28&29.47&75.85&58.35&29.68&74.03\\
		DiT   \textit{{\scriptsize (CVPR'23)}} &49.05&29.57&74.53&62.22&\textbf{29.90}&\textbf{71.45}\\
		EMamba \textit{{\scriptsize (Arxiv'24)}}   & 45.36   & 28.76  & 89.08 & 64.61   & 29.08   & 86.26 \\
		ZigMa  \textit{{\scriptsize (Arxiv'24)}}           & 42.18   & 29.06 & 82.74 & 63.73    & 29.37  & 81.49  \\
		ViM  \textit{{\scriptsize (Arxiv'24)}}           &  41.49  &  29.38 & 76.34 & 60.31   & 29.38  & 80.30 \\
		VMamba \textit{{\scriptsize (ICML'24)}}      & 51.32   &  29.60  & 74.38 & 62.52   & 29.34  & 81.58 \\
		\midrule
		DiffMa \textit{{\scriptsize (ours)}} & \textbf{56.59}   &  \textbf{29.76} &\textbf{71.90} & \textbf{69.60}   &  29.40 & 79.96 \\
		\bottomrule
	\end{tabularx}
\end{table*}

\section{Experiment}
In this section, we first introduce the dataset and training details. We then conduct a series of extensive experiments to compare DiffMa with baseline models, including Diffusion models based on CNNs and ViTs. In particular, we replace our Mamba block with several newly proposed visual Mamba structures to verify the effectiveness of our designed architecture.

\subsection{Dataset}
We use the SynthRAD2023 dataset~\cite{thummerer2023synthrad2023}, which includes CT, CBCT, and MRI scans of brain and pelvic radiotherapy patients from three university medical centers in the Netherlands. We exclude the CBCT data, focusing instead on the CT and MRI data. Importantly, the primary task of the SynthRAD2023 Grand Challenge is to generate CT rather than MRI. However, considering the critical importance of MRI in clinical diagnoses and the complexity of its image features, in this work, we endeavor to address the more challenging task of CT to MRI conversion.

\begin{figure}[t]
	\centering
	\includegraphics[width=0.472\textwidth]{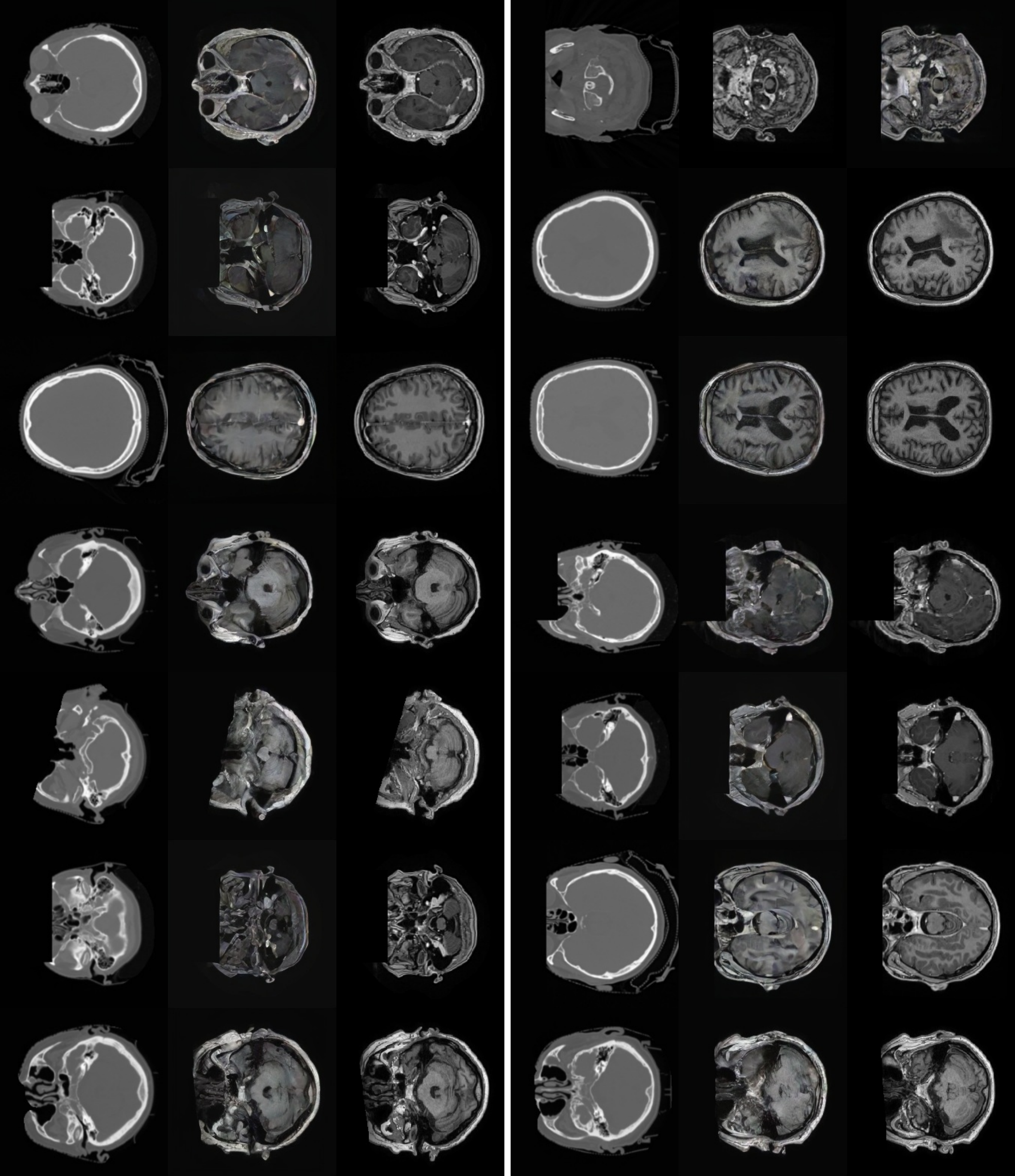}
	\caption{\textbf{Visualizations of 14 brain CT to MRI conversion pairs from the SynthRAD2023 dataset.} Among each pair, \textit{Left} is the inputed CT, \textit{Middle} is the generated MRI and \textit{Right} is the ground truth MRI. \textit{Zoom in for a better view.}}
	\label{fig:pelvis}
\end{figure}

\subsection{Training strategies}
The DiffMa model was trained using image with resolution of $224\times224$. 
The last linear layer, as shown on the \textit{left} side of Figure~\ref{fig:framework}, was initialized to $0$, with standard weight initialization applied to the remaining layers. 
The AdamW optimizer~\cite{kingma2014adam, loshchilov2017decoupled} was employed with a learning rate of $1e-4$ and without weight decay. Data augmentation was not employed, diverging from common practices in natural image generation. We did not tune decay/warm-up schedules, learning rate, or Adam $\beta_1/\beta_2$. Training incorporated the use of an Exponential Moving Average (EMA)~\cite{tarvainen2017mean} with a decay rate of 0.9999.
All experimental results presented employ EMA model. 

\begin{figure}[t]
	\centering
	\includegraphics[width=0.472\textwidth]{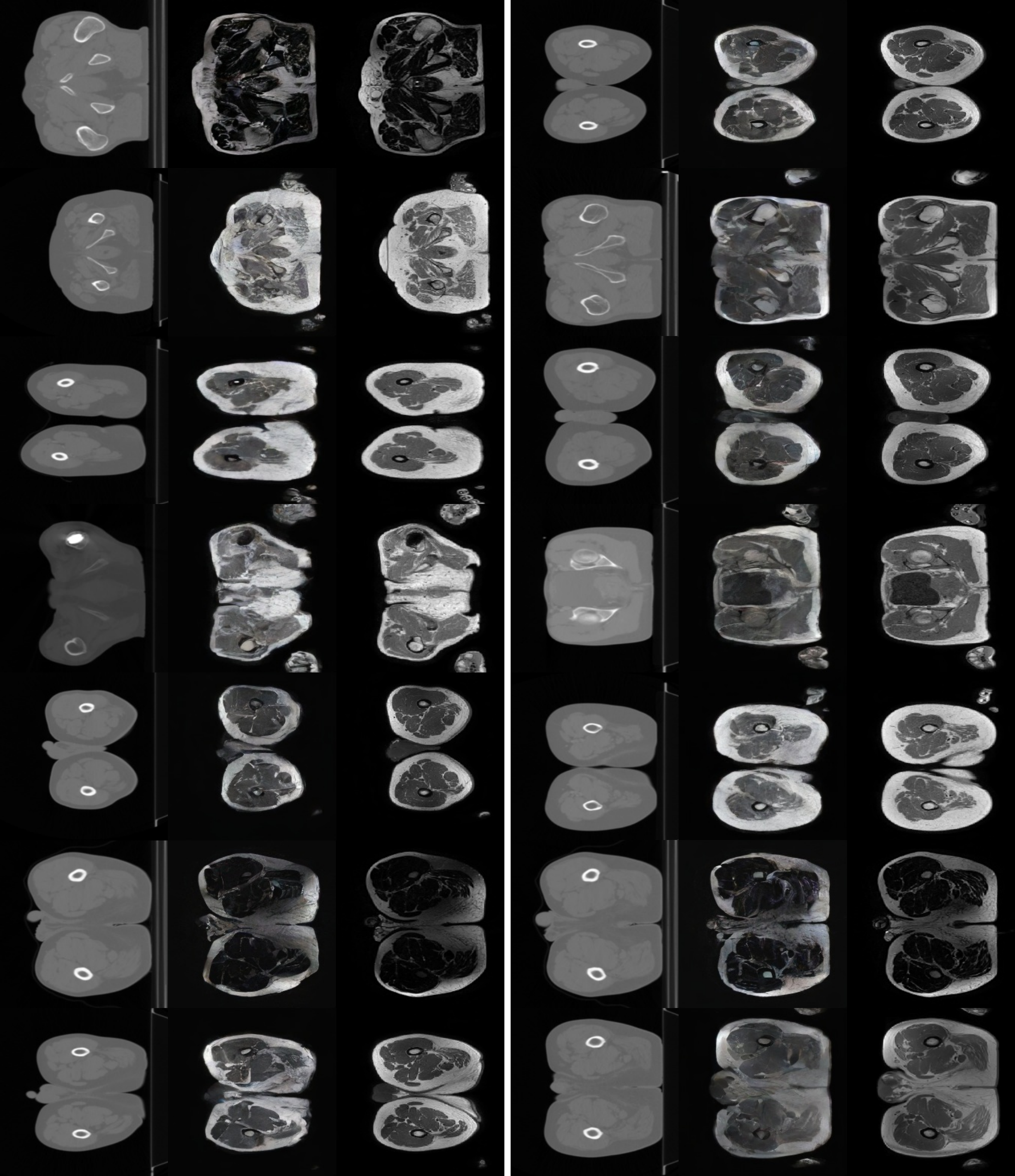}
	\caption{\textbf{Visualizations of 14 pelvis CT to MRI conversion pairs from the SynthRAD2023 dataset.} Among each pair, \textit{Left} is the inputed CT, \textit{Middle} is the generated MRI and \textit{Right} is the ground truth MRI. \textit{Zoom in for a better view.}}
	\label{fig:brain}
\end{figure}

\begin{figure*}[htbp]
	\centering
	\includegraphics[width=\textwidth]{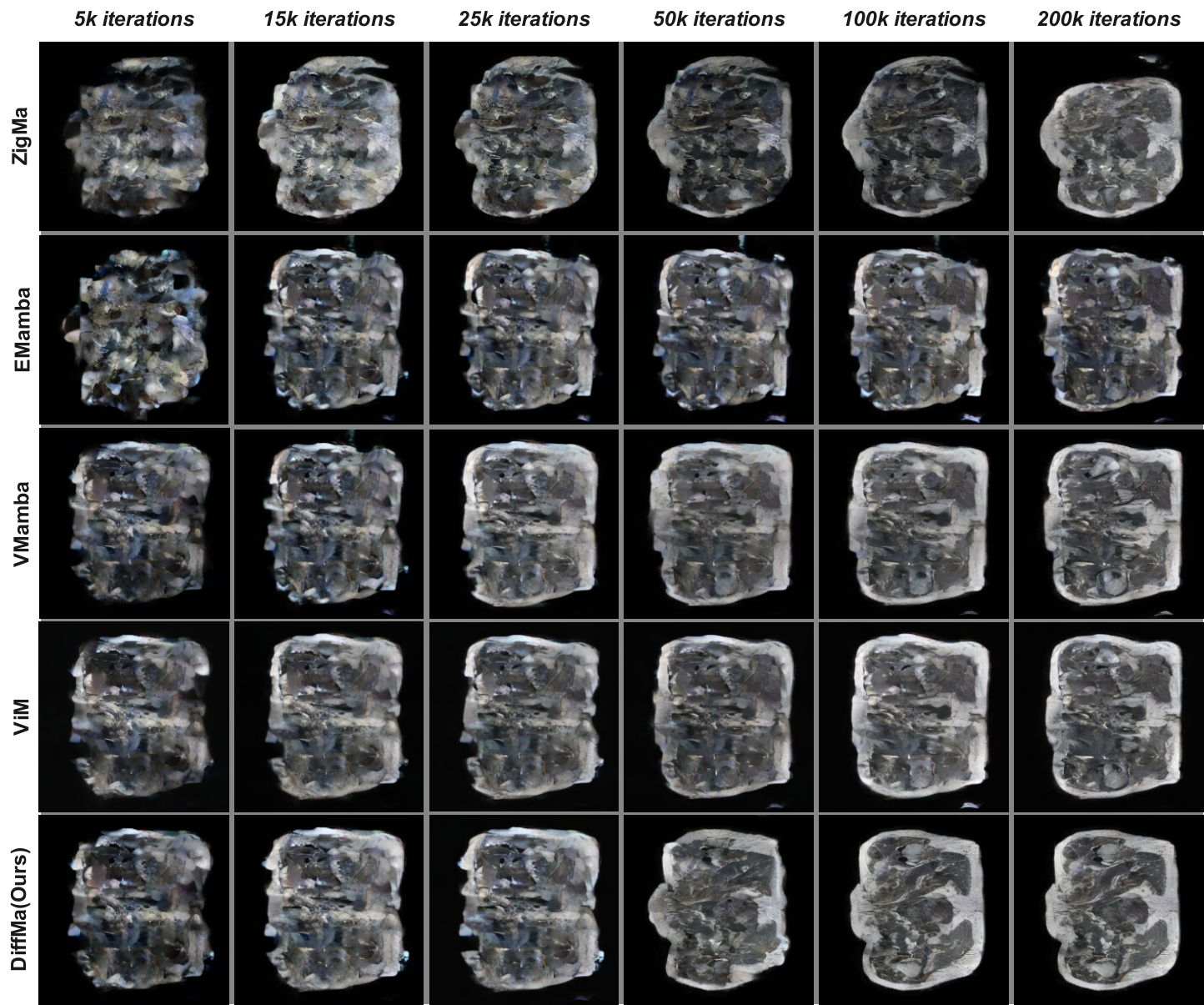}
	\caption{\textbf{Visualizations of generating MRI using different methods}. Our method DiffMa produced the best image quality with fewer iterations.}
	\label{fig:ex_iter_compare}
\end{figure*}

We used an off-the-shelf pre-trained VAE model from Stable Diffusion~\cite{rombach2022high}, downsampled by a factor of 8.
All experiments in this section are conducted within the latent space produced by the VAE encoder, with the sampling outcomes subsequently reconstructed to pixel form via the VAE decoder.
The linear variance schedule $t$ has a maximum value $1000$.
Most of our experiments are conducted using Nvidia RTX 3090 Ti GPUs.
For sampling, we employ ODE sampling due to speed considerations.

\subsection{Results}
We present the visualization results of our framework on the CT to MRI conversion task in Figures~\ref{fig:pelvis} and \ref{fig:brain}, illustrating the precision and fidelity of DiffMa.
Additionally, we compared our method with CNN-base Diffusion method LDM~\cite{rombach2022high}, ViT-base Diffusion method DiT~\cite{peebles2023scalable}, and several recently proposed Mamba block structures, including VMamba~\cite{liu2024vmamba}, EMamba~\cite{pei2024efficientvmamba}, ZigMa~\cite{hu2024zigma} and ViM~\cite{zhu2024vision}, and the results are presented in Tables~\ref{table:compare}. Under the same training epochs and model parameter count, our method demonstrated significant improvements. It is important to note that although EMamba, ViM and VMamba are commonly used for classification and segmentation tasks, we adapted them for generation tasks by replacing their blocks with ours. The model weights are available at \textit{\textcolor{red}{https://huggingface.co/ZhenbinWang/DiffMa/tree/main}}.

The evaluation metrics include the Structural Similarity Index (SSIM), Peak Signal-to-Noise Ratio (PSNR), and Mean Squared Error (MSE). PSNR and MSE quantify the pixel-level differences between two images without considering visual perception. However, SSIM takes into account the brightness, contrast, and structure of the image, aligning more closely with the human eye's assessment of image quality. From Table~\ref{table:compare}, the following conclusions can be drawn: 1) At the same number of iterations, Diffusion models employing ViTs and Mamba architectures demonstrate a notable superiority in SSIM metrics over those utilizing CNNs architectures. This enhancement is likely attributable to the constrained receptive fields of CNNs, in contrast to the comprehensive global modeling capabilities of both ViTs and Mamba architectures;
2) Diffusion models based on ViTs and CNNs perform well on PSNR and MSE metrics, outperforming several Mamba-based architectures. Notably, they achieve the best performance on the brain dataset, although this requires sacrificing the global receptive field or entails high computational costs. Despite these trade-offs, the results reveal pixel-level biases in most vision Mamba architectures. Our method DiffMa addresses these shortcomings to some extent, particularly evident in the pelvis dataset, where DiffMa surpasses CNN and ViT-based architectures in both PSNR and MSE metrics.

To further emphasize the strengths of the proposed method, the image generation capabilities of EMA models from various Mamba blocks were compared, maintaining the same parameter count and iteration numbers. As depicted in Figure~\ref{fig:ex_iter_compare}, the results demonstrate that the proposed method DiffMa achieves superior performance with fewer iterations compared to other methods.

\section{Conclusion}
We introduce Diffusion Mamba (DiffMa), a Mamba-based diffusion model designed for CT to MRI conversion task. DiffMa integrates the advantages of State-Space models, including linear complexity and a global receptive field, into medical image generation. We address key issues of cross-sequence attention and spatial continuity by introducing soft mask and Spiral-Scan scheme. 
Experimental results confirm that our designed block outperforms other vision mamba blocks in CT to MRI conversion task. 
We hope that our endeavor can inspire further exploration in the Mamba-based diffusion network design.

\bibliography{aaai23}

\end{document}